\documentclass[alpha-refs]{nbdt-article}

\usepackage{siunitx}
\usepackage{ccfonts}
\usepackage{hyperref}

\papertype{Original Article}
\paperfield{Journal Section}

\title{Artificial Intelligence is Algorithmic Mimicry: \\ \LARGE Why artificial ``agents'' are not (and won't be) proper agents}

\author[1,2,3]{Johannes Jaeger}

\affil[1]{Department of Philosophy, University of Vienna, Austria}
\affil[2]{Complexity Science Hub (CSH) Vienna, Austria}
\affil[3]{Ronin Institute, U.S.A.}

\corraddress{Johannes Jaeger, Department of Philosophy, University of Vienna, Universit\"{a}tsstra{\ss}e 7, 1010 Vienna, Austria}
\corremail{johannes\_jaeger@univie.ac.at, yoginho@ronininstitute.org}

\fundinginfo{John Templeton Foundation, Grant Number: 62581}

\runningauthor{Johannes Jaeger}

\begin{document}

\maketitle

\begin{abstract}
\small What is the prospect of developing artificial general intelligence (AGI)?\\ I investigate this question by systematically comparing living and algorithmic systems, with a special focus on the notion of ``agency.''\\ There are three fundamental differences to consider:\\ (1)~Living systems are {\em autopoietic}, that is, self-manufacturing, and therefore able to set their own intrinsic goals, while algorithms exist in a computational environment with target functions that are both provided by an external agent.\\ (2)~Living systems are {\em embodied} in the sense that there is no separation between their symbolic and physical aspects, while algorithms run on computational architectures that maximally isolate software from hardware.\\ (3)~Living systems experience a {\em large world}, in which most problems are ill-defined (and not all definable), while algorithms exist in a small world, in which all problems are well-defined.\\ These three differences imply that living and algorithmic systems have very different capabilities and limitations.\\ In particular, it is extremely unlikely that true AGI (beyond mere mimicry) can be developed in the current algorithmic framework of AI research.\\ Consequently, discussions about the proper development and deployment of algorithmic tools should be shaped around the dangers and opportunities of current narrow AI, not the extremely unlikely prospect of the emergence of true agency in artificial systems.

\end{abstract}

\pagebreak

\section{Introduction}
\label{sec_introduction}

There has been much discussion about the prospect of artificial general intelligence (AGI). This debate was triggered by recent advances in the field of artificial intelligence (AI), including the invention of \textit{transformer models}, derived from multi-layered (deep) recurrent neural networks, and a massive scaling-up in \textit{machine-learning (ML)} research. This has resulted in the arrival of \textit{large language models (LLMs)}, trained on enormous datasets of very broad range (\cite{vaswani2017attn, qiu2020pretrained, bender2020climbing, bender2021stochasticparrots, han2021pretrained, shanahan2023langmodels}). Such models (\textit{e.g.} BERT, \cite{devlin2019bert}; the GPT series, \cite{radford2019langlearn}; or LaMDA, \cite{thoppilan2022lambda}) are capable of astonishing feats of inference in the domain of language, exhibiting text-generation and conversational capabilities that have impressed (and even overwhelmed) many a limited human observer, expert or otherwise. Further progress is expected from multimodal versions of such models, capable of handling not just language, but images, video, and music (see, for example, \cite{takagi2022diffmodels, huang2023langmodels}).

Opinions on the potential of these algorithms range from them being mere ``stochastic parrots'' ``haphazardly stitching together sequences of linguistic forms'' obtained from training data ``without any reference to meaning" (\cite{bender2021stochasticparrots}) to current LLMs already exhibiting some ``sparks of AGI'' (\cite{bubeck2023sparks}). The latter view originates with corporate AI engineers directly working on these models, who justify their claims with the fact that these large inference machines can seemingly ``solve novel and difficult tasks'' in diverse areas ``without needing any special prompting'' (\textit{ibid.}). In another, heavily publicized example, ChatGPT has been alleged to be proficient at chemistry, despite ``nobody having programmed it to learn chemistry.''\footnote{This claim did not originate with researchers, but comes from U.S. Senator Chris Murphy who posted it on social media (\href{https://twitter.com/ChrisMurphyCT/status/1640186536825061376?s$=$20}{https://twitter.com/ChrisMurphyCT/status/1640186536825061376?s$=$20}). It triggered numerous critical replies from experts (see, for example, \href{https://www.thedailybeast.com/senator-chris-murphys-chatgpt-tweet-is-proof-lawmakers-arent-ready-for-the-ai-boom}{https://www.thedailybeast.com/senator-chris-murphys-chatgpt-tweet-is-proof-lawmakers-arent-ready-for-the-ai-boom}).} Similar anecdotes abound, illustrating the widespread impression among researchers and the general public alike that LLMs may have acquired some capabilities akin to general intelligence.

One common speculation is that the extremely high-dimensional nature of the weight matrix (and thus the configuration space) of an LLM allows for the emergence of ``higher-level'' abilities, such as the detection of meaning without external referents (see, for example, \cite{piantadosi2022}). It is a short step from there to claim that such higher-level phenomena could potentially include things such as agency, intelligence, or even consciousness. Due to our own innate cognitive limitations, no human being is able to understand the detailed mechanisms by which a model with billions of weights detects correlations that are of much higher dimensionality than the ones our brains could possibly pick up or process. This leads to predictive capabilities in LLMs that are mind-boggling and seemingly miraculous to any human observer. Humans have evolved to see agency everywhere in nature (\cite{kuhnen2020agencydet}). We are easily impressed (and fooled) that way. But does this imply that true higher-order effects could arise inside such a complicated algorithmic system? Could LLMs really become true agents, or even conscious, one day?

Here, I focus on the concept of ``agency,'' and conclude that the probability of an LLM ever achieving --- as opposed to merely simulating, emulating, or mimicking --- something analogous to natural or organismic agency is almost infinitesimally small. Achieving true artificial agency, and thus artificial cognition or consciousness, would require major conceptual breakthroughs in algorithm design as well as radical innovations in materials and computational architecture of a kind that are not currently on the horizon. In fact, I argue that no purely algorithmic (syntactic) computational system is capable of true embodied agency even if it is embedded in robotic hardware. It simply makes no sense to say that algorithms ``act'' in the world like living beings do. As a consequence, they do not ``think'' like humans either. In fact, they do not possess anything like animal or human cognition, meaning that they cannot become conscious (as humans are) no matter what improvements in computing power and training data await us in the future. Although they may emulate specific cognitive tasks to a degree where they outdo human performance, they will not surpass us in general intelligence, whatever that may mean. Artificial superintelligence is an unrealistic myth. Algorithms are and remain what they have always been: machines --- automated tools for computation. We had better treat them as such. 

My argument places me very firmly at the ``stochastic parrots'' end of the debate. The truth does not always lie in the middle. To defend this seemingly extreme position, I will use arguments from biology that are surprisingly absent from the current discussion about artificial intelligence (AI). They complement powerful arguments against AGI based on (computational) linguistics, which show that LLMs cannot extract meaning from natural language despite appearing to emulate it with surprising accuracy and flexibility (\cite{bender2020climbing, bender2021stochasticparrots}). 

In my argument, I have to deal with two fundamental problems. The first is rooted in the fact that concepts have very different meanings in different disciplines. Here, I focus on why the term ``AI agent'' is a gross misnomer, as algorithms have no agency in the biological sense of the term. Similar arguments can be made about the term ``artificial intelligence'' itself, which describes a research discipline which is not concerned with ``intelligence'' at all. We need better terminology to describe what is going on. In particular, I suggest supplanting ``artificial intelligence'' with ``\textit{algorithmic mimicry}'' which circumscribes much better what the field is doing. The term ``AI agent'' should simply be avoided. ``Algorithm'' suffices completely.

The second problem is that the current debate ignores the fact that organismic and algorithmic systems are built on architectures that are radically and fundamentally different. While computers are designed on the principle of a maximum \textit{separation} or \textit{fractionability} of hardware and software, organismic agency requires a completely different organization based on the maximum \textit{integration} of physical and symbolic\footnote{It is hard to avoid the terms ``symbol'' and ``symbolic'' here. We mostly use ``symbol'' in a broad sense as ``something that stands for something else by reason of a relation'' (\cite{pattee2012laws}). This is not to be confused with the more specific use of the term in AI research, where ``symbolic'' indicates a method based on high-level (human-readable) problem representations (in contrast to subsymbolic methods such as connectionist networks).} (\textit{i.e.}, code-related) aspects (see section \ref{sec_embodiment}, and \cite{rosen1991life, pattee2012laws, barbieri2015codebiology}). In this particular sense, organisms and computers seem to be exact opposites of each other. This means that algorithms can only imitate (\textit{i.e.}, simulate, emulate, or mimic), but not truly reproduce or represent higher-level phenomena such as agency, which is exclusive to living matter and its peculiar organization (\cite{rosen1991life, kauffman2000investigations, moreno2015autonomy}). This is the main reason I think that AGI is not possible in the current algorithmic framework of AI research: the architecture of this framework is too flat --- it does not allow for the kind of hierarchical circularity required for natural agency --- no matter how large or sophisticated its models. There is no spark of AGI anywhere to be seen. In fact, I predict that, if AGI is ever created, it will come out of a biological laboratory (see section \ref{sec_largeworld}).

I proceed by giving a brief introduction to what I mean by ``general intelligence'' and ``agency.'' I then use these concepts to show three fundamental differences between organisms and algorithms (which are a kind of machine; see section \ref{sec_embodiment}, for a detailed explanation). The first difference is \textit{autopoiesis}, the ability of organisms to self-manufacture, which essentially grounds their natural agency. The second is \textit{embodiment}, the tight integration of ``hardware'' and ``software,'' which governs an organism’s interactions with its environment and renders the quality of those interactions fundamentally different from those of a machine. The third is the fact that organisms live in \textit{a large world}, while algorithms exist in a small world, no matter how high-dimensional their parameter space or how enormous the dataset they have been trained on (\cite{savage1954statistics}). In a large world, information is typically scarce, ambiguous, and often misleading, which makes it difficult to identify relevant problems and characterize them with any precision. In contrast, a small world is entirely made of data that are explicit and well-defined. Based on these three fundamental differences, I argue that it is unreasonable and dangerous to mistake the virtual for the real world. Unfortunately, this is exactly what is obscuring our debates about algorithmic mimicry and its real-world implications at the moment.

\section{General Intelligence and Agency}
\label{sec_genint_agency}

To rigorously assess the prospect of AGI, I need to first define what I mean by ``\textit{general intelligence}.'' Debates about this topic all too often reflect differences in operational definitions of concepts. One particular problem is that researchers in the field of algorithmic mimicry tend to have an overly simplistic conception of ``intelligence.'' They consider it merely a matter of problem solving, often formalized as optimization in the context of some kind of general problem-solving framework, such as the one originally proposed by  \cite{newell1972problem}. This kind of approach is problematic in itself, since it comes up against the problem of relevance, \textit{i.e.}, how to formally define a real-world problem in the first place (see section~\ref{sec_largeworld}, and  \cite{vervaeke2012relevance, vervaeke2013relevance}). More importantly, it is far too narrow to embrace the everyday meaning of ``general intelligence.'' To capture this broader conception, I take the definition to include the following minimal set of characteristics (\cite{roitblat2020algorithms}):
\begin{itemize}
	\item reasoning (especially inference-making),
	\item problem solving,
	\item learning,
	\item using common-sense knowledge,
	\item autonomously defining and adjusting goals,
	\item dealing with ambiguity and ill-defined situations, and
	\item creating new representations of the knowledge acquired.
\end{itemize}
I have argued elsewhere that all of these characteristics of general intelligence are essential for an organism to get to know its world (\cite{roli2022organisms}). But only the first three can be properly formalized (and only explicitly propositional forms of learning at that; see, for example, \cite{polanyi1958personal}). Algorithmic mimicry is exclusively concerned with these formalizable characteristics, but cannot deal with the other four that are equally important. This generates a number of largely intractable problems. I note that there has been very little progress on these matters in the 70 years since the goal of general intelligence was first posited at the Dartmouth Summer Research Project in 1956 (\textit{e.g.}, \cite{mccarthy1969ai, dreyfus1972what, dennett1984cogwheels, dreyfus1992whatstill, cantwellsmith2019ai, roitblat2020algorithms, roli2022organisms}), and I see no realistic prospect of this state of affairs changing any time soon. AGI is not an imminent possibility, if it is achievable at all.

Not enough people realize that the problems preventing AGI are of a philosophical rather than technological or practical nature (\cite{roli2022organisms}): they all have to do with the fact that algorithms and the environment they exist in, by definition, are purely syntactic constructs without the possibility of acquiring semantic referents in the physical ``outside'' world (section \ref{sec_embodiment}). Consider, for instance, that ``common sense'' cannot be precisely defined in any purely syntactic and formal (and hence general) sense, as it applies only in the contingent, real-world context of the semantics of a living social organism. Similarly, algorithms cannot freely set their own intrinsic goals because they remain bound by their instructions, input data, and computational environment which are provided externally (see below and section~\ref{sec_autopoiesis}). Neither can they deal with ambiguous, ill-defined problems (section~\ref{sec_largeworld}). There are no double-entendres in a purely syntactic construct. 

Last but not least, algorithms do not have the capacity to create new frames or representations of their ``knowledge'' since they exist in a completely formalized (small) world, where everything is already well-defined (section~\ref{sec_largeworld}). This world, in its entirety, consists of the algorithm’s instructions and computational environment, plus training (and later input) data that have to be properly formatted with regard to some predefined objective (no matter how broadly defined). In other words, the algorithm’s model of the world \textit{is} its world. The algorithm cannot switch frames because there is only one frame: its complete \textit{digital ontology}. Nothing and everything is relevant at the same time in such a situation.

The digital ontology of an algorithm includes all correlations read out of data, no matter how high-dimensional and subtle. These correlations are not emergent in the same physical sense a candle flame or a living being is. They are not really higher-level phenomena, because they are fully precoded in the data, although perhaps hidden and undetectable for limited human observers. Furthermore, such correlations cannot be true semantic representations, because they have no real meaning about anything beyond the algorithm’s limited digital ontology. Instead, they are implicit in purely syntactic programs and data structures  (\cite{bender2020climbing, piantadosi2022}). If semantics are inferred, they are only derived from internal relations between symbols, as when LLMs haphazardly manage to arrive at the ``meaning'' of a term from clustering of terms into word classes in their underlying vector space (\textit{ibid.}).

According to the definition of ``general intelligence'' given above, whether or not an algorithm can acquire AGI hinges crucially (among other things) on whether or not it can set its own goals (\cite{roitblat2020algorithms}). Being able to set intrinsic goals, in turn, is a basic and essential property of what I call the ``\textit{natural agency}'' of living beings (see, for example, \cite{barandiaran2009agency, kauffman2000investigations, moreno2015autonomy, walsh2015organisms, roli2022organisms}). Organisms, from bacteria to humans, are natural agents because they can define and pursue their own goals.

There is nothing mysterious or unscientific about natural agency, and it does not require any cognition or intention, as we shall see (\textit{cf.} \cite{barandiaran2009agency}). Instead, natural agency is simply grounded in the ability of all organisms to self-manufacture (\cite{hofmeyr2021cellmodel}) --- in the fact that they are \textit{embodied autopoietic systems} (see section~\ref{sec_autopoiesis}). An autopoietic system is organized in a way that enables it to produce and maintain itself by continuously fabricating and assembling its own physical components (\cite{varela1974autopoiesis, varela1979autonomy, maturana1980autopoiesis}). For example, a cell self-manufactures by producing macromolecular components (proteins, nucleic acids, lipids) through metabolism, by enabling the functional folding and assembly of those components (into protein complexes or membranes, for instance) through the maintenance of a specific cellular milieu, and by sustaining this milieu via the regulated transmembrane transport of nutrients, electrolytes, and waste products (\cite{hofmeyr2021cellmodel}). More generally, the primary intrinsic goal of any organism is to continue existing, to go on self-manufacturing, to keep alive\footnote{As always in biology, there are exceptions to this rule. Most cases where organisms sacrifice themselves as individuals can be explained by the higher-level perspective of inclusive fitness: they do it for the benefit of their offspring, relatives, or conspecifics. A curious case where this does not apply is the tragic phenomenon of human suicide. But even here, it takes considerable effort (and the very peculiar nature of the human condition) to subvert the innate survival instinct.}. It is what it does, naturally, and nobody told it to do so. This is the constitutive dimension of being an autonomous living agent (\cite{moreno2015autonomy}). 

But it is not sufficient. To stay alive, an organism also needs to be able to initiate actions that are aligned with its particular environment (\cite{moreno2015autonomy, walsh2015organisms}). In other words, survival requires well-adapted, goal-oriented behavior. Such behavior can (and often does) result from evolution by natural selection. It requires internal predictive models of the world, understood very broadly as processes or structures within the organism that function in relation to projected outcomes of its actions  (\cite{rosen1985anticipatory, louie2017anticip}). These ``models'' need not be representational, nor do they need to be based on cognition or intention. Often, they are actualized by simple evolved mechanisms. Think of a bacterium able to discern toxins from nutrients and ``going for'' food by swimming up a nutrient gradient. Even the simplest organism is an \textit{ anticipatory system} (\textit{ibid.}). All such systems use predictive models to pursue their intrinsic goals. This is what enables organisms to act on their own behalf  (\cite{kauffman2000investigations}). Or refuse to act, for that matter. It represents the interactive dimension of being an autonomous living agent (\cite{moreno2015autonomy}).

Algorithms are only superficially like that. Granted, they can have purposes, can even possess a certain degree of autonomy (the ability to select from a range of tasks or objective functions without direct human intervention), and can implement models of the physical world. Accordingly, the Encyclopedia Britannica defines a software ``agent'' as ``a computer program that performs various actions continuously and autonomously on behalf of an individual or an organization.''\footnote{Retrieved at \href{https://www.britannica.com/technology/software-agent}{https://www.britannica.com/technology/software-agent} on Apr 21, 2023.} Think of self-driving cars, or any other autonomous robot, for example. The important part of the definition in the present context is that the algorithm’s ``actions'' and ``autonomy'' always manifest themselves ``on behalf of an individual or organization,'' which means they ultimately come from a true natural agent, a human programmer, systems designer, or data scientist, who provides instructions, objectives, labeled data, and computational environment. An algorithm’s goals are ultimately always imposed from outside. An algorithm always acts on behalf of an agent, or it does not act at all. It has no natural agency.

Sometimes, we are deceived into thinking this is not the case, because the coding of the tasks and objectives of an algorithm happens in a very implicit and indirect manner, as in current ML research with its large datasets and diversified broad-range ``unsupervised'' learning strategies. And yet, even the ``smartest'' LLM model cannot set its own goals. In fact, it never will. LLMs, by definition, are programmed to do a specific task, even if it is as abstract and general as ``finding high-dimensional correlations in very large and intricate datasets in order to complete phrases formulated in some language.'' An LLM’s autonomy is inflexible, predetermined externally, and thus remains much more circumscribed than that of a true natural agent.

All of this means that an algorithm builds its model of the world (its ``knowledge'' if you will) in a very different way from a living being (\cite{roli2022organisms}). In fact, the two have almost nothing in common. First, an algorithm’s model of the world is always built with regard to an outside agent’s goals and, second, the algorithm does not really \textit{have} a model of the world but rather \textit{is} some kind of model of something. For example, LLMs are models of natural language (as their name implies), while the software of a self-driving car is a model of that part of physical reality that represents traffic. It bears repeating: algorithms are (and will always remain) \textit{tools for real agents}. They are not agents themselves.

To better understand why this is the case, let us now look at the three fundamental differences between organisms and algorithms.

\section{Autopoiesis}
\label{sec_autopoiesis}

As we have seen, the primary goal of an organismic agent is to remain alive, while machines have no such intrinsic drive towards self-preservation (section~\ref{sec_genint_agency}). Without this drive, there can be no natural agency, and without agency no true general intelligence. But what about the likelihood of such an intrinsic drive arising in an algorithmic system in the future? Could an algorithm eventually become a true agent, say, above a certain level of computational complexity? Would that mean it has become alive? To evaluate these questions, we must first better understand what life is or, more precisely, what distinguishes living from non-living matter.

The difference is one of organization rather than composition. Organisms are composed of chemical elements that are also common in non-living systems and, like everything else in the universe, they must obey the fundamental laws of physics. What really sets living matter apart is not what it is made of, but the way in which the physico-chemical processes that are its components influence and constrain each other’s dynamical behavior (\cite{rosen1991life, juarrero1999dynamics, kauffman2000investigations, deacon2011incomplete, pattee2012laws, moreno2015autonomy, montevil2015constraints, juarrero2023context}). In mathematical terms, such \textit{constraints} are called boundary conditions on the underlying flow. They limit the degrees of freedom that a particular component process can actualize. They narrow its range of possible dynamical behaviors. They restrict what it can do. Interestingly, life is all about constraints. In the words of Terrence \cite{deacon2011incomplete}: living systems are \textit{less} than the sum of their parts!  

Life can only exist far from chemical equilibrium. As far-from-equilibrium systems, organisms must be thermodynamically open, constantly exchanging materials and energy with their environment. More specifically, organisms are \textit{dissipative systems} (\cite{prigogine1973dissipative, nicolis1977selforg, prigogine1984order}): they deplete naturally occurring gradients of free energy (\textit{i.e.}, entropy) at the maximal possible rate. While dissipative systems include non-living self-organizing processes, such as hurricanes, eddies, and candle flames, organisms go one step further. They use physical work driven by a free energy gradient to generate internal constraints which, in turn, channel the underlying physico-chemical processes in specific directions that keep the whole process going (\cite{kauffman2000investigations, deacon2011incomplete, montevil2015constraints}). Put simply, organisms keep themselves alive by using constraints to transform and build further constraints. 

The constructive dynamic of building constraints upon constraints is what leads to \textit{autopoiesis}, the ability of the organism to self-manufacture (\textit{cf.} section~\ref{sec_genint_agency}). Autopoiesis arises when each constraint within the system is not only generated by but also generates at least one other constraint (\cite{montevil2015constraints, mossio2016organization}). The constraints that constitute the autopoietic \textit{organization} of an organism collectively produce each other. Think of the set of enzymes present in some cell. They collectively produce themselves through their role in metabolism and gene regulation. This defining property of a living system is called \textit{organizational closure} (\cite{piaget1967biocon, moreno2015autonomy}). Let us emphasize again that organizational closure requires thermodynamic openness. It is only possible in a system that remains far from equilibrium. 

As an illustrative example, consider the self-manufacturing organization of a free-living cell (\cite{hofmeyr2021cellmodel}): metabolism produces nucleotides, proteins, and lipids (among many other things) through macromolecular synthesis. The resulting macromolecules, in turn, acquire their specific functional conformation (or assemble into higher-level structures such as ribosomes, protein complexes, and membranes) given the tightly regulated internal milieu of the cell. Finally, this milieu itself needs to be constantly monitored and maintained through the regulation of transmembrane transport, which requires an assembled membrane system and functional protein transporters. It is easy to see how all three aspects of this process constrain, but also rely on each other for their own existence. This dynamic is what is meant by ``constraints building upon constraints.'' Through the dialectic interrelation of synthesis, milieu, and transport, the cell is able to self-manufacture, and ultimately reproduce by cell division.

In other words: to stay alive (to be an autopoietic system), an organism must maintain organizational closure over time, throughout its entire life cycle. On top of that, the organism must pass on its organization to its offspring, across generations, if it is to be evolvable (\cite{saborido2011heredity, mossio2020heredity}). This is called \textit{organizational continuity}  (\cite{difrisco2020diachronic}). It is what underlies the individuality of living systems (\textit{ibid.}). In addition, organizational continuity is an essential requirement for (open-ended) evolution of physical (embodied) systems by natural selection (\cite{roli2022organisms, jaeger2024fourth}). Finally, and most importantly in the current context: it is what gives the organism a certain degree of \textit{autonomy} from its environment, based on its capacity for \textit{self-determination} (\cite{deacon2011incomplete, moreno2015autonomy, mossio2017teleology}).

Self-determination is possible because the particular direction that the constructive dynamic of autopoietic constraint generation is taking is not exhaustively determined by the laws of physics that govern the underlying processes that are being constrained  (\cite{deacon2011incomplete, pattee2012laws}). Instead, constraint generation follows its own inner logic grounded in the hierarchical and self-referential interrelations of processes in a living system, and how they support each other and, at the same time, restrict each other’s range of possible behaviors. 

This dynamic at the level of constraints is historically contingent. It is evolutionary because it not only depends on the particular environmental context of an organism (section~\ref{sec_embodiment}), but is also \textit{dynamically presupposed} by earlier organized states \textit{within} the living system (and its ancestors; \cite{bickhard2000autonomy, mossio2017teleology, difrisco2020diachronic}). In this sense, the process of constraint generation is fundamentally unpredictable, that is, \textit{radically emergent}, to any outside observer (\cite{kauffman2000investigations, roli2022organisms}). Living beings can behave and evolve in ways that are not foreseeable, even in principle. This is what it means to have \textit{autonomy}. Moreover, the behavior of an organism originates within its own organization. This is what it means to have \textit{agency}. Autonomous agents (and systems that contain them) do not break any laws of physics, but are not reducible to physics and chemistry either, since their behavior is not predictable by physical law.

Can we reproduce autopoietic organization in an algorithmic system? Can we generate \textit{artificial autopoiesis}? At some level, definitely yes: there is no reason to assume it is impossible to emulate autopoietic processes with algorithmic mimicry. In fact, we have known for a long time that self-producing dynamics can be generated in cellular automata (\cite{vonneumann1966automaton, hofmeyr2018causation}). Similarly, the hierarchically circular dynamics of closure and autopoiesis can be captured with \textit{abstract rewrite systems}, for instance, computational models that are based on $\lambda$-calculus or related formalisms in which operations are allowed to redefine the rules of the program (\cite{fontana1994arrival, fontana1996barrier, mossio2009computable}). Thus, at first sight, there is no reason to assume that the dynamics of living systems cannot be fully captured by algorithmic simulation.

It is worth noting, however, that none of the formalisms or hardware implementations currently employed in the field of algorithmic mimicry are based on the autopoietic principles described here. The hierarchical self-referentiality of closure represents a kind of \textit{organizational complexity} that is not the same as the \textit{computational complexity} of cybernetic feedback and recursive computing (\cite{rosen1991life, louie2009ml}). The former has to do with the way the physical component processes of a living system interrelate to mutually support and restrict each other (constraints building upon constraints), while the latter is about the reducibility (\textit{i.e.}, compressibility) of the underlying processes themselves. An algorithm is irreducible, if we cannot skip any of its steps and still obtain the same behavior. It can produce surprising outcomes since we cannot follow each individual step of the calculation, but its boundary conditions (instructions and computational environment) remain predefined and fixed. An organism, in contrast, is irreducible in terms of its organization: it produces unpredictable behavior because of the way it constructs its own constraints. Autopoiesis is irreducible because it is the property of a whole, intact physical system. It should not be difficult to see that the two kinds of complexity are fundamentally different. 

Contemporary approaches to algorithmic mimicry, such as LLMs and other models based on recurrent neural nets (section~\ref{sec_introduction}), are rich in computational complexity, but lack the organizational complexity required for autopoiesis. To put it another way: even if ``deep'' or multi-layered, their architecture is too flat to truly capture the kind of hierarchical circularity required for autopoiesis. Since autopoiesis is a prerequisite for self-determination, it is safe to conclude that current AI algorithms will not exhibit true agency (as in ``natural agency'') any time soon, no matter how many parameters a model contains, how many network levels it features, or how large and complex the datasets used to train it. Agency is not a matter of size or scale. It is a matter of organization.

But what if artificial autopoiesis \textit{will} be developed? Will an autopoietic simulation truly represent a living system in the sense of capturing all its essential characteristics, as well as its full behavioral and evolutionary potential? Will such a system qualify as being alive in some sense? I argue that this is extremely unlikely. First of all, there are convincing arguments --- based on the mathematical theory of categories --- which suggest that any algorithmic simulation of a living system must necessarily remain incomplete  (\cite{rosen1991life, louie2009ml, hofmeyr2021cellmodel}). It cannot fully capture the behavioral and evolutionary potential of an organism due to the collectively impredicative nature of the latter (\textit{ibid.}). I shall not go into that rather technical discussion here (but see  \cite{roli2022organisms}). Instead, I will focus on another limitation of algorithms that is often overlooked: it lies in the very different ways in which syntactic codes relate to the physical world in machines and living systems. This is the problem of embodiment, which is, in essence, a generalized variant of the symbol grounding problem in cognitive science (\cite{harnad1990grounding}).

\section{Embodiment}
\label{sec_embodiment}

Digital computers are designed with the cleanest possible separation between hardware and software in mind (\cite{rosen1991life, pattee2012laws}). The rationale for this is practical: you want to be able to perform as many different automated computation tasks as possible on the same hardware. This is why a large majority of modern computers share the same basic architecture, which is derived from a general abstract model of computation: a \textit{universal Turing machine} (\cite{turing1936computable, davis2001engines}). This mathematical machine is a maximally flexible calculating device. It is conjectured to provide a general model of effective computation, that is, a general model of the kind of computation a human can perform by rote (\cite{church1936unsolvable, turing1936computable}). In this sense, Turing’s machine can perform \textit{any} kind of automated calculation, run \textit{any} kind of software, implement \textit{any} possible algorithm\footnote{There is one important caveat here: Turing machines have no clocks that measure real time, only state transitions. This is why many real-time computing systems are not strictly Turing machines. But this fact does not affect our argument.}. 

I define an algorithm as a sequence of precisely defined logical or mathematical operations that reliably performs some computation, typically, to solve a specific problem. Because I am not concerned with the practical aspects of problem-solving here, I interpret ``algorithm'' in a broader sense than is usually done in the theory of computation. I do not require the computation to halt, that is, the number of steps involved to be finite. Also: ``sequence'' does not imply strict sequentiality. Computational threads can run concurrently, as long as the order of their interactions is well-defined. It is even possible to add a non-deterministic element by allowing for a set of possible operations to be applied probabilistically at each step. No matter how we specify the term, an algorithm represents a purely automatic procedure that requires no agency (\cite{rosen1991life}).

A Turing machine consists of an infinite tape with symbols drawn from a finite alphabet, and a reading head that can be in a finite number of different states (\cite{turing1936computable}). The head reads the symbol at the present position on the tape, then performs an operation that is determined by this symbol and the current state the head is in (or, if the machine is non-deterministic, it draws an operation probabilistically from a given set). The machine then either halts, or writes an output symbol to the tape while remaining where it is or moving to the left or right to continue its operation. The symbols on the tape are the data the machine is operating on. The transition table that determines (a set of) operations for each symbol and each state defines the algorithm it implements. This table can be stored inside the head, but it can also be read from the tape. The latter results in what is called a \textit{stored-program computer}: not only the data, but also the algorithm it executes are on the tape (\cite{davis2001engines}). Apart from the infinite sequential tape, which is replaced by finite random-access memory, this is the basic architecture of today’s digital computers. Thus, computers are imperfect real-world approximations of a (stored-program) Turing machine.

A Turing machine with an infinite tape can implement any arbitrary sequence of symbols, and hence the complete set of codable algorithms. These algorithms are its software, while the reading head is the hardware. It is in this sense that this idealized model of a real-world computer is a \textit{universal} machine. Turing’s model gives us a precise definition of what is meant by ``computation.'' A process is computable if it can run on a universal Turing machine (\cite{turing1936computable, copeland2020churchturing}). 

The model also yields a technical definition of a ``machine,'' which does not coincide with the way we use the term in everyday language. Instead, a machine in the sense of Turing is a formal or natural system (a mathematical or physical automaton) that corresponds to a universal Turing machine (\cite{rosen1991life}). Some physicists have speculated that \textit{all} natural processes must be machines in this sense (\cite{gandy1980church, deutsch1984quantum, deutsch1997fabric, lloyd2007programming}; recently reviewed in \cite{piccinini2021physcomp}). In this view, the whole world literally \textit{is} an automaton: any process that is real must be representable in algorithmic terms, including all living processes. If we subscribe to this pancomputationalist stance, AI algorithms \textit{must} have the capacity to become true agents, to become alive, to become conscious, if only we manage to capture the right set of computational properties of a living system. Pancomputationalism is behind the enthusiastic claims about AGI cited in section~\ref{sec_introduction}.

Unfortunately, pancomputationalism is fundamentally missing the point. Turing’s model of computation is concerned with the utterly human cognitive activity of ``efficiently performing a calculation,'' not with physics (\cite{copeland2020churchturing}). Mistaking the two is so common it has its own name: \textit{the equivalence fallacy} (\textit{ibid.}). What Turing’s model \textit{does} is enable engineers to build powerful computers, which automate the rote activity of calculating. Our digital computers are tools that approximate the functionality of universal machines, capable of performing any kind of effective calculation. This is only possible \textit{because} of the strict software-hardware separation in the model and in real-world digital computers. However, this separation also means that algorithms exist strictly in the symbolic realm of software, isolated from the physical details of the hardware they run on. In this sense, running a simulation on a digital computer is an almost completely ``physics-free'' activity (\cite{pattee2012laws}).

Let us step back and reflect once more on the claim that all physical mechanisms must be Turing-computable. This claim goes far beyond the trivially true statement that we can \textit{approximate} physical processes by simulating them in a computer (\cite{pattee2012laws}). It implies that every physical process must have some symbolic content (otherwise, it would not be strictly equivalent to an algorithm). But this is problematic because of the ``aboutness'' of symbols: they are about something else; they must have a referent outside their own strictly syntactic domain (\cite{deacon2011incomplete}). Furthermore, the meaning of a word in a natural language crucially depends on the communicative intent of the speaker and the interpretation by the listener (\cite{bender2020climbing}). Similarly, the purpose of a calculation is that of the natural agent performing it, no matter whether it is performed in the agent’s head or automated on a digital computer (section~\ref{sec_genint_agency}). There is no way around it: symbols with meaning are tightly tethered to the existence of living agents in the physical world. Physical mechanisms that are not machines (that are not designed by an agent for some purpose) have no intrinsic symbolic content. We can impute meaning on them by simulating them, but they are \textit{not} machines in the sense of being the same as a Turing machine. Most physical mechanisms do not perform any computation, only our simulations of them do.

Thus, machines (in the technical sense used here) are the small subset of mechanisms which have a purpose imposed on them by an external agent. Most of them are \textit{our} machines, machines that humans have designed, constructed, or programmed. Once we know its purpose, it is possible to describe a machine in algorithmic terms, that is, in terms of its function (\cite{kauffman1971articulation, rosen1991life}). Functional (symbolic) and mechanistic (physical) descriptions complement each other and, in the case of a machine, map onto each other in a straightforward manner: it is generally possible to localize specific functions to particular physical parts of the machine’s mechanism. 

All algorithms are mechanisms if actualized in the context of a physical automaton: their operation is precisely defined by orchestrated sequences of accurately localizable cause-and-effect relations between the parts that constitute the machine (\cite{bechtel2011mechanism, nicholson2012mechanism, glennan2017mechphil}). Unlike digital computers, most mechanical machines only actualize a small set of algorithms that serve a specific function, \textit{e.g.}, solving a particular problem. A lawn mower implements the rote procedure of mowing a lawn. A car serves our need for transportation.

Like machines, organisms are natural systems with a purpose (see sections~\ref{sec_genint_agency} \& \ref{sec_autopoiesis}). This means they are amenable to functional descriptions, i.e., we can differentiate them into symbolic and physical aspects, which can (if one is so inclined) be analogized to the software and the hardware of a computer (\textit{e.g.}, \cite{rosen1991life, pattee2012laws, barbieri2015codebiology}). Unlike machines, though, the goals of an organism originate \textit{within} the organization of the living system itself, and are not imposed from outside (\cite{nicholson2013organism, nicholson2014machine}). This is reflected in a fundamental difference between biological organization and computational architecture: organisms lack any separation of ``hardware" and ``software''. Symbolic and physical aspects of a living system are intimately intermingled. They may be distinguishable conceptually, but cannot be disentangled from each other like they can in a computer (\cite{pattee2012laws}).

Take, for example, the central relation between genome and protein sequence in a living cell. This relation can be viewed as symbolic in the sense that it is mediated by the genetic code, where specific DNA codons ``stand for'' particular amino acids\footnote{I understand concepts like ``code'' and ``symbol'' to be explicitly context-dependent and historically contingent here, not suggesting any universal (but arbitrarily coded) ``book of life'' that ``writes out'' (and thus determines) the living form (see also \cite{kay2000book, foxkeller2010mirage}, for criticisms of language metaphors in genetics and molecular biology).}. Yet, the meaning of a genomic sequence is only acquired through the physical processes that express a gene to produce a functional protein in a given context. This requires a set of appropriate enzymes to be present, whose primary structure is also coded in genomic sequence. However, expressed proteins only attain their functional three-dimensional conformation after folding occurs in the tightly regulated biochemical milieu of the cell (\cite{pattee2012laws, barbieri2015codebiology, hofmeyr2018causation, hofmeyr2021cellmodel}). This last step is \textit{not} symbolically coded in the genome. Folding is purely physical, requiring a cellular milieu that is tightly regulated at the level of the whole cell \cite{hofmeyr2021cellmodel}.

Here, the difference to machines should become immediately obvious: the decoding process (the ``software'' --- often mislabeled ``genetic program'') produces enzymes that (through folding in a particular milieu) constitute the physical ``hardware'' of the cell, which are in turn required to replicate and activate the ``software.'' While genomic sequences are often analogized with the tape of a Turing machine, and transcription/translation enzymes with the reading head interpreting those symbolic sequences, the analogy breaks down when we consider that the reading head itself is generated by the decoding process, and that its components require whole-cell regulatory processes outside the genome (plus the laws of thermodynamics) to take on their functional forms. In other words, in a cell, the ``hardware'' \textit{is} the ``software'' \textit{is} the ``hardware,'' and so on. They are distinct but inseparable aspects of the \textit{same} overall process, similar to the difference between constraints and underlying flux described in section~\ref{sec_autopoiesis}.

A symbolic system that works in this integrated way is no longer formally equivalent to a Turing machine (\cite{rosen1991life, louie2017anticip, hofmeyr2021cellmodel}). It literally \textit{is} what it \textit{does}. The continual mutual interconversion between the symbolic and the physical is a fundamental organizational principle of living systems, and a type of embodiment that results in a deep and direct embedding of the symbolic aspects of the organism in its physical context. Howard Pattee has called this interconnectedness between symbol and matter \textit{semantic closure}\footnote{The use of this term by Pattee is a bit confusing, since it is not the same concept as that first proposed by Alfred Tarski to denote a theory or language that can adequately express its own semantic concepts (\cite{tarski1956logic, priest1984semclos}).} (\cite{pattee2012laws}). It is basically a version of von Neumann's (1966) self-reproducing automaton that is not isolated from, but compatible with, the laws of physics. It is what enables autopoiesis, self-manufacture, in the real world. Compare this to the almost complete separation of software and hardware in ``physics-free'' computational architectures. In this sense, organisms are the exact opposite of machines like digital computers and the algorithms they run. 

While a living system literally is what it does in the physical world, the relationship between an algorithm isolated in its purely syntactic domain (section~\ref{sec_genint_agency}) and its physical surroundings is much more complicated and much less direct. This difference in the quality of embodiment does not depend on the details of the computation: it applies to \textit{anything} that runs on a digital stored-program computer, no matter what kind of algorithmic mimicry is performed, and no matter how complicated the algorithm or how comprehensive the dataset it was trained on. Not even embedding software in robotic hardware removes the disconnect, irrespective of what kind of peripherals are involved, as long as the algorithm is still bound by the principles of digital computer architecture. An algorithm has no possibility of transcending its digital ontology. By definition, it exists in a closed world that it cannot escape (see section~\ref{sec_largeworld}). If it has an effect on the physical world (beyond its consumption of electricity and taking up space), it either achieves it through influencing the behavior of the external agent that is using it, or (in robotics) through effectors that are part of the hardware it runs on. Both of these ways of interacting require additional components and interfaces based on rules that must be provided externally. They are not the direct product of the software, as enzymes (and other ``hardware'' components) are in living systems.

To overcome this fundamental limitation would require an algorithm to generate its own hardware, according to its own intrinsic goals and specifications. This poses several rather serious challenges. First of all, intrinsic goals would require a computational architecture that enables true (natural) agency. We do not know at this point what such an architecture would look like (see section~\ref{sec_autopoiesis}). Second, direct embodiment necessitates a completely different computational paradigm, which achieves the greatest possible universality without any strict separation of hardware and software. Efforts towards neuromorphic computing can be seen as a tiny first step in this direction, but we are still very far from accurately emulating the development and function of nervous systems in animals, not even to mention self-manufacture (see, for example, \cite{schuman2017neuromorphic, schuman2022opportunities}). And finally, autonomous hardware evolution will only be possible with much more flexible and configurable materials than we currently have available\footnote{The best we currently have in terms of configurable hardware are the above-mentioned neuromorphic nets, and field-programmable gate arrays. The latter are electronic circuits that can be dynamically reconfigured by software (within predefined limits) but are nowhere near being generated by the algorithm according to its own requirements.} (\textit{cf.} \cite{moreno2005agency, nicholson2019cell}). None of these daunting challenges are likely to be overcome any time soon, if they can be overcome at all. Therefore, AI algorithms will remain confined to the symbolic realm for the foreseeable future, relying on human beings to mediate their effect on the world. Taken together, this suggests that unaligned autonomous AGI, running free outside its preconfigured computational environment, is an extremely unlikely scenario at this point. Rather than worrying about the potential rise of superintelligent machines, we had better pay close attention to the pernicious effects narrow AI can have on our own, very human, behavior.

\section{Large Worlds}
\label{sec_largeworld}

In the previous two sections, we have seen that organisms are organized and embodied in a way that is very different from machines, including all current attempts at algorithmic mimicry. As a consequence, organisms and algorithms exist in vastly different worlds. Following \cite{savage1954statistics}, I shall call these worlds ``large'' and ``small.''

The \textit{small world} of an algorithm is purely syntactic and symbolic (just like the algorithm itself), maximally isolated from the messy, ambiguous, and often misleading semantics of physical reality (section~\ref{sec_embodiment}). It is a formal construct encompassing the algorithm’s own code, its formatted data (training as well as input), and the computational architecture it is embedded in (hardware design, operating system, and language environment). On the one hand, the algorithm has access to its world in its entirety --- nothing ever remains obscure or hidden. On the other hand, it is enclosed in its small world for good --- there is no way for it to go beyond its given digital ontology, to change its frame of reference. If the algorithm ``perceives'' the physical world, it is through an encoded interface with hardware sensors; if it is embedded in robotic hardware, it acts through a decoding interface with hardware effectors. We have seen in section~\ref{sec_embodiment} that the interaction of an organism with its surroundings is much more immediate. Unlike the algorithm, whose interfaces must be provided by an external (human) agent, a living being generates the structures it needs to interact with the physical world from within its own organization.

In an algorithm’s closed small world, there is only one frame of reference that includes its entire ``universe,'' a finite (though potentially astronomically vast) set of variables and rules defining their relations, all explicitly expressed in terms of a specific formalism. All accessible features of such a world are predetermined (however implicitly and indirectly) by this formalism, and get prioritized in regard to a target function (or a range of target functions). Again, both formalism and target function(s) are externally imposed by some (human) agent. In such a small world (no matter how vast), everything and nothing is relevant at the same time. Every problem the algorithm could possibly encounter is clearly defined: it can be assigned an initial program state, a target state (producing some desired output), and a discrete and finite search space containing possible computations that connect the two (\cite{newell1972problem}). Note that being well-defined does not mean all these problems are solvable, as some will turn out to be computationally complex, and hence intractable (as defined in section~\ref{sec_autopoiesis}).

Often, we consider organisms to function in a similar way. But this is what philosophers call a \textit{category error}. Computationalist approaches to cognition, for example, tend to see the mind as operating in a manner that is analogous to the software of a computer --- isolated from physical reality through the ``hardware'' making up our bodies, in particular, our channels of perception. In this view, sensory organs and nervous systems work like hardware sensors that encode input we receive from the physical world. The human mind, like an algorithm, appears to exist in a completely symbolic and syntactic (and thus small) world. 

However, the differences in organization and embodiment discussed in sections~\ref{sec_autopoiesis} \& \ref{sec_embodiment} suggest an alternative scenario. Since organisms are self-manufacturing and have no software-hardware distinction, cognition itself must be treated as an autopoietic and embodied phenomenon (see, for example, \cite{varela1991embmind, thompson2010mindlife}). Even though there may be some encoding going on in sensory perception, and there may be abstract symbolic representations involved, at heart, our interaction with the physical world is much more intimate and immediate than that of an algorithm. This is because our hardware is made out of software and vice versa (section~\ref{sec_embodiment}). You may have sensors and effectors, but unlike the robot, these are made directly from your ``software,'' that is, the coding processes involved in macromolecular synthesis. It all becomes clearer when we focus on simpler organisms without nervous systems, where the connection to the physical world is much less convoluted than in cognitive agents such as humans.

Consider, again, a single free-living cell. Its autopoietic nature is characterized by two defining properties. First and foremost, it is a true agent, able to set and pursue its own goals (section~\ref{sec_autopoiesis}). It needs no target function. Its primary goal is to remain alive. To achieve this goal, it must self-manufacture, which means it is constantly forced to invest physical work into maintaining a set of constraints that keeps it far from equilibrium. Hans \cite{jonas1966phenomenon} called this the \textit{thermodynamic predicament}. On the one hand, life is precarious that way. On the other, its autonomy gives the cell a certain degree of self-determination (\cite{mossio2017teleology}). \cite{jonas1966phenomenon} calls this flip side of the coin our \textit{needful freedom}. In contrast, an algorithm does not (in fact, cannot) expend energy to persist. It can be stored, erased, and reloaded indefinitely without affecting its performance. It has no control --- nor any kind of subjective awareness --- of its own existence.

Second, any living being is an open but bounded system, with its boundaries one of the vital constraints that it must continue generating from within itself (section~\ref{sec_autopoiesis}). These boundaries mediate the interactions the cell can have with its physical surroundings, which make up its experience of the world. Note that ``experience'' in this sense does not imply any cognitive capabilities or mental representations, which a single cell does not have. 

What it does imply, however, is that a cell has some kind of ``point of view,'' a peculiar frame of reference, which it creates itself from within its own organization and through which it experiences the world. In contrast, an algorithm always ``sees'' everything in its small world. It has a view ``from nowhere.'' The reference frame of an organism is shaped by evolution: its interactions (and the structures that mediate them) must be reasonably well adapted to the environment to ensure the continued existence of the cell. Taken together, all this means that the experience of any living being is limited by necessity. This is exacerbated by the fact that organisms are tiny (in fact, almost infinitesimal in size) compared to their environment. For a living being, there is no ``god’s eye view'' (\cite{giere2006perspectivism, wimsatt2007reveng, massimi2022perspreal}). Its experience of the world will always be biased and partial.

In conclusion, living beings, unlike algorithms, live in a \textit{large world}, a world far beyond their limited grasp (\cite{stanford2010exceeding}). This is not only a vast and ancient world (small worlds can be like that too), but a world in which most problems are \textit{not} well-defined in the sense introduced above. Thus, by definition, a large world is not formalized. And it is probably not formalizable, meaning that a limited observer would never be able to express every possible problem in terms of specific initial and end states, plus a discrete and finite search space. There is always a \textit{semantic residual} of phenomena that remain vague and mysterious. This is the main difference between the worlds of organisms and algorithms. Information, in a large world, is always scarce, often ambiguous, and sometimes outright misleading. This means that an organism, if it is to solve problems at all, must first define them, turn cryptic semantics into clear-cut syntax, which involves the dilemma of having to identify and tackle those aspects of the world that are relevant to survival. 

This is a dilemma, because it is generally not solvable algorithmically. There is no precisely circumscribed search space and, if there were such a thing, this would only lead us into an infinite definitional regress: in order to delimit a search space precisely we have to identify the relevant aspects of the problem, which defines an optimization problem that requires a well-defined search space, and so on. To avoid this regress means to overcome the \textit{problem of relevance} (\cite{vervaeke2012relevance}), which is a generalized version of what researchers in the field of algorithmic mimicry know as the \textit{frame problem} (reviewed in \cite{shanahan2016frameprob}). Even the simplest organism can deal with it. A bacterium, for example, has evolved mechanisms to distinguish between chemicals that are nutrients versus those that are toxins in its environment. In contrast, it is a notoriously intractable problem for AI (see, for example, \cite{mccarthy1969ai, dreyfus1972what, dennett1984cogwheels, dreyfus1992whatstill, cantwellsmith2019ai, roitblat2020algorithms}). And here is why: apart from the fact that nothing is intrinsically relevant to an algorithm because it is not an autopoietic agent (section~\ref{sec_autopoiesis}), there is only one frame in a small world, and thus no frame problem to be solved. In other words, the problem of relevance simply does not exist in the world of algorithmic mimicry.  

This is not a technological problem, but a philosophical one. There will be no technological solution to it in the current framework of algorithmic mimicry, no matter how complicated and powerful our methods for inference, our training data, and our hardware architecture will get. Expressed in colloquial terms, an algorithm cannot want or need anything, because it is not alive. In contrast, organisms are essentially driven by desire and impermanence. These are the foundations for meaning and relevance. Life is where the transition from matter to mattering occurs (\cite{roli2022organisms}). The precariousness of life is what motivates the actions of a living being, which are carried out with the primary goal of staying alive. As long as we don’t generate ``artificial intelligences'' that exist in a large world, there will be no AGI. As already stated in section~\ref{sec_introduction}, such a system is more likely to emerge from a biological laboratory than any current effort in artificial mimicry.

\section{Conclusions}
\label{sec_conclusions}

In this paper, I highlight three basic differences between living and algorithmic systems. First, organisms are self-manufacturing (autopoietic) physical systems, which can identify, set, and pursue their own intrinsic goals, while algorithms are purely symbolic machines that are dependent on a suitable computational environment and target functions that must be provided by some external agent. Second, the prevalent computational architecture of today maximizes the isolation of software from hardware, meaning that interactions with the physical world require externally provided sensors and effectors, while no software-hardware distinction exists in organisms, which are embodied in a way that enables more immediate exchange with their physical surroundings. Finally, algorithms exist in a small world, in which all possible problems are well-defined, whereas organisms live in a large world, where most problems are ill-defined (and some are probably not properly definable). All of this goes to show just how different living systems are from algorithms.

This means that algorithms and living beings have very different capabilities and limitations. Surprisingly, this fact is often overlooked in discussions that compare algorithmic mimicry with natural general intelligence. It is true that algorithms outperform humans in many tasks. Those tasks are typically well-defined but hard for us to solve, especially if they involve large amounts of calculation, require substantial working memory, and/or involve high-dimensional search spaces. Familiar examples are strategic games like chess or go, complicated scheduling or planning tasks, intricate mathematical proofs that require a large number of steps, or (in the case of LLMs) filling in the blanks of a text based on massive amounts of correlations between words and/or phrases in a training dataset. Because humans are cognitively rather limited in these areas, never having evolved the ability to solve complicated problems of this kind, we are easily impressed by the algorithms’ performance. Sometimes, this causes us to lose perspective, to attribute capabilities to these machines that they cannot possibly have because of their architectural limitations. 

To paraphrase Robert \cite{rosen1991life}, algorithms represent automatic procedures that require no thought, no interpretation, no improvisation, and no intrinsic agency or creativity. And this, as explained in section~\ref{sec_embodiment}, is exactly the purpose of the Church-Turing theory of computation: it is a model of what humans can achieve by rote calculation. Only much later, after the computer came into widespread use --- after it became the defining high-tech of our age --- did it become widely accepted to apply the theory of computation as a general model of physics or cognition. 

However, most physical processes have no intrinsic symbolic content, and are therefore not ``computation'' in Church and Turing’s sense of the term. It takes a living agent to impute symbolic meaning onto reality, either by simulating physical systems (to understand or control them), or by building machine-artifacts that materialize certain algorithmic processes. 

Similarly, in the domain of cognition, processes of rote calculation are only a tiny fraction of what an animal or human brain does. Other neural and cognitive processes may be meaningfully emulated by computation, but they are not necessarily computational in nature. In fact, brains did not evolve to be calculating machines in the sense of Turing at all. Instead, they arose as a means of coordinating an animal’s embodied sensorimotor control in the context of a large and complex world. In the words of Paul \cite{cisek2019resynthesizing}, on p.~2270: ``[T]he evolutionary history of the nervous system is essentially a history of the continuous extension of such control further and further into the world." Brains enable an organism to tackle the problem of relevance in ever more involved situations. I have established in section~\ref{sec_largeworld} that this problem is not tractable algorithmically.  

To summarize: my argument shows that, even though computation can and does occur in physical and cognitive systems, it is a category error to consider such systems purely in terms of Turing-style computation. Computationalism is not wrong per se, it is a useful tool for simulating certain cognitive processes, but we must learn to recognize its limited domain of application. Ignoring these limitations turns a blind eye to the fact that algorithmic systems may outperform humans at complicated tasks of learning, inference-making, and problem-solving in well-defined situations, while they cannot compete with us in situations that demand choosing and setting intrinsic goals, using situated common sense, dealing with ill-defined problems and ambiguous or misleading information, and/or changing frame of reference. I, and others, have argued elsewhere that \textit{all} of these abilities constitute essential aspects of general intelligence  (\cite{roitblat2020algorithms, roli2022organisms}).

Therefore, thinking of algorithmic mimicry (including LLMs) in terms of ``agency,'' ``cognition,'' ``thought,'' ``understanding,'' or ``intelligence'' means using the wrong categories\footnote{Alison Gopnik has written an excellent and very accessible column on precisely this topic: \href{https://www.wsj.com/articles/what-ai-still-doesnt-know-how-to-do-11657891316}{https://www.wsj.com/articles/what-ai-still-doesnt-know-how-to-do-11657891316}.}, because none of these capabilities, as originally defined in biology, are achievable in today’s algorithmic framework of AI research. This also implies that AGI, or conscious algorithmic ``agents,'' are not a realistic imminent possibility. 

At the heart of this conceptual confusion lies a distinction that is rarely recognized. When we talk about the emergence of novel capabilities in algorithmic and living systems, we use two fundamentally different notions of ``emergence,'' based on two completely distinct definitions of ``complexity'' (see section~\ref{sec_autopoiesis}). 

In an algorithmic context, we are dealing with \textit{computational complexity}, which captures the irreducibility of certain computational processes, and the effort it takes to calculate them in a realistic time frame (\textit{e.g.}, \cite{aaronson2013compcomplex, dean2021compcomplex}). Roughly, we have to perform certain computations step-by-step to arrive at the desired target state, but this may not be possible to do faster than in real time. In other words, there is no short-cut, for instance, through general dynamical laws that allow predicting the outcome without performing the entire calculation. In such systems, emergent properties are defined as features that are computationally irreducible. They are unexpected (\textit{i.e.}, emergent) only in the sense that we cannot easily predict them in advance. Philosophers call this \textit{weak emergence} (\cite{bedau1997weakemergence, humphreys2015emergence}).

In contrast, the kind of complexity that characterizes living systems is not computational but \textit{organizational complexity}. Autopoietic systems contain hierarchical cycles, which lie at the heart of their ability to self-manufacture. These cycles also implement organizational closure and underlie the tight integration of hardware (physical) and software (symbolic aspects) in such systems  (\cite{rosen1991life, deacon2011incomplete, pattee2012laws, montevil2015constraints}). The kind of emergence we get in this case is the emergence of new rules that govern the dynamics of the system. This is how new levels of organization arise \cite{wimsatt2007reveng, deacon2011incomplete}. ``Agency,'' ``cognition,'' ``thought,'' ``understanding,'' and ``intelligence'' inhabit these higher levels of organization. The computational architecture of contemporary algorithmic mimicry has no way to allow such new levels to emerge. Its complexity, as breathtaking as it may be, remains strictly confined to the computational kind. I repeat once more: the emergence of AGI is not a matter of size or scale, it is a matter of organization.

To conclude, I will state again how important it is to use appropriate language when talking about algorithms such as LLMs. These systems do not ``conceptualize,'' ``conceive,'' or ``create.'' They ``count,'' ``calculate,'' and ``compute.'' The term ``artificial intelligence'' itself is a gross misnomer: the work in this field, as it currently stands, has nothing to do with natural intelligence. I suggest calling it \textit{algorithmic mimicry} instead, which makes its nature explicit and helps to avoid category errors such as the ones described above. Or, when mimicry is well done and useful to human agents (which it often is), we could call it IA: \textit{intelligence augmentation}. Algorithms are tools, admittedly more complex than a hammer, but still tools that we can use, if we choose to, for boosting our own cognitive capabilities. An AI ``agent'' is never an agent on its own.

My argument supports those who see the dangers of AI not in the possibility of AGI (which is not a real possibility right now), but in dangerous and deceptive applications of narrow algorithmic mimicry. LLMs are and remain in this narrow category, no matter how astonishing their feats of imitation. The \textit{problem of alignment} is not one of adjusting ourselves to the presence of superior entities. Quite the contrary, we must recognize these algorithms for what they are: powerful tools to be adjusted to \textit{our} human needs (see, for example, \cite{werthner2022dighum}). This is an urgent and tremendous practical problem that will have to be solved using societal and political means. It is also, to a large degree, a problem of design: algorithms must be clearly distinguishable from real agents because the two are not at all similar in kind. As Daniel Dennett puts it: ``[c]ounterfeit money has been seen as vandalism against society ever since money has existed. Punishments included the death penalty and being drawn and quartered. Counterfeit people is at least as serious\footnote{Dennett was quoted in a feature that New York magazine ran on computational linguist Emily Bender called ``You are not a parrot:'' \href{https://nymag.com/intelligencer/article/ai-artificial-intelligence-chatbots-emily-m-bender.html}{https://nymag.com/intelligencer/article/ai-artificial-intelligence-chatbots-emily-m-bender.html}.}. I agree. Regulation is urgent and indispensable. Responsibility, in the end, lies with us human agents. \textit{We} are the ones with the agency. Why voluntarily delegate it to a machine that has none? 

\section*{Acknowledgements}
Paul Poledna introduced me to the term ``IA (Intelligence Augmentation).'' The incentive to write this paper arose from discussions within our project ``\textit{Pushing the Boundaries}'' which is led by myself and Tarja Knuuttila, and is hosted by the Department of Philosophy of the University of Vienna. I thank all participants for their support and inspiration. Specifically, I would like to thank Erich Prem, Tarja Knuuttila, Andrea Loettgers, Paul Poledna, Kevin Purkhauser, and Mortz Kriegleder for comments on the manuscript. Last but not least, I would like to extend a very special ``thank you'' to the late Brian Goodwin, and my good friend Jannie Hofmeyr, who are the ones who got me thinking about these kinds of topics in the first place.

\section*{Conflict of Interest}
The author has no conflicts of interest.

\bibliography{algorithmic_mimicry}

\end{document}